\documentclass{article}
\usepackage{spconf,amsmath,graphicx}
\usepackage{stfloats}
\usepackage[a4paper,margin=1in]{geometry}
\usepackage{amsmath,amssymb,amsfonts} 
\usepackage[utf8]{inputenc}           
\usepackage[draft]{hyperref}
\usepackage{amsmath}
\usepackage{amssymb}
\usepackage{graphicx}
\usepackage{float}
\usepackage{booktabs}
\usepackage{multirow}
\usepackage{longtable}
\usepackage{subcaption}
\usepackage{adjustbox}
\usepackage{caption}
\usepackage{tikz}
\usetikzlibrary{arrows.meta,positioning}
\usepackage{pgfgantt}
\usepackage{gensymb}
\usepackage{datetime2}
\usepackage{ifthen}
\usepackage{enumitem}
\usepackage{float}  
\usepackage{subcaption}

\usepackage{enumitem}
\setlist{nosep, leftmargin=14pt}

\usepackage{mwe} 

\title{The Good, the Bad, and the Brittle: Benchmarking Robustness and Generalisation of Histopathology Foundation Models}
\name{Dhyey Yajnik \qquad Amina Asif \qquad Fayyaz Minhas\thanks{Corresponding author: \href{mailto:fayyaz.minhas@warwick.ac.uk}{fayyaz.minhas@warwick.ac.uk}}}
\address{Predictive Systems in Biomedicine Lab, Tissue Image Analytics Centre,\\
Department of Computer Science, University of Warwick, United Kingdom}

\begin{document}
%
\maketitle
\begin{abstract}
\emph{How robust and generalisable are pathology foundation models and have their scaling limites been reached?}
We benchmarked twelve pathology foundation models (PFMs) and ResNet baselines using our Robustness Evaluation and Enhancement Toolbox (REET) across eleven clinically realistic perturbations and a dissimilarity-driven Non-Redundant K-fold validation (NR-Kfold) protocol. We introduce a Perturbation Performance Index (PPI) to summarise accuracy trends under controlled perturbation sweeps and analyse robustness scaling with parameter count. We show that PFMs consistently outperform CNNs in both robustness and domain generalisation, yet model scaling shows diminishing returns: mid-sized models such (UNI2/Virchow-2 etc.) achieve comparable or greater resilience than larger systems. NR-Kfold analysis further reveals systematic accuracy loss and increased variability when training-test similarity is broken, underscoring the need for explicit distribution-shift evaluation. These findings suggest that the next generation of pathology foundation models must prioritise data quality, multimodality information and domain alignment over parameter count to achieve genuine clinical reliability.

\end{abstract}
%
%
\section{Introduction}

Deep learning has transformed computational pathology, enabling automated analysis of large-scale histology data and discovery of diagnostic and prognostic biomarkers.  
However, clinical reliability demands more than high test-set accuracy: models must remain stable under staining variation, scanner bias, and imaging artefacts that often cause sharp performance degradation~\cite{MedCentDiff, reetoolbox}.  
Beyond such perturbations, robustness to broader distributional differences between centres,  populations, subtypes, etc., i.e., \emph{domain generalisation}, is equally critical.  
These include covariate, prior, and class-conditional shifts arising from variations in scanners, staining protocols, or tissue composition across sites~\cite{jahanifar2025domain}.  

While early convolutional networks achieved strong in-domain performance but failed across centres, recent pathology foundation models (PFMs) trained on massive within-domain corpora using self-supervised objectives have markedly improved transferability by learning stain- and tissue-invariant representations~\cite{campanella25_foundational_model_comparison}.  
Yet, despite their scale and sophistication, PFMs remain fragile under realistic perturbations and unseen domains, underscoring the need for systematic and quantitative robustness assessment.

To address this, we previously developed the Robustness Evaluation and Enhancement Toolbox (REET)~\cite{reetoolbox}, a pathology-specific framework that generates clinically realistic perturbations (e.g., stain, blur, compression, geometry) and probes model stability through stochastic and adversarial optimisation.  
While REET has revealed important robustness limitations in conventional histopathology models, three key questions remain open for foundation-scale architectures:  
\begin{enumerate}
    \item \textbf{Robustness to perturbations:} How resilient are PFMs to clinically realistic image variations?  
    \item \textbf{Domain-shift generalisation:} How do they perform under dissimilar data distributions?  
    \item \textbf{Scaling effect:} Does model size continue to improve robustness, or have gains saturated?  
\end{enumerate}

\section{Materials and Methods}
To address these research questions, we implemented a unified robustness and generalisation framework based on the Robustness Evaluation and Enhancement Toolbox (REET)~\cite{reetoolbox}, evaluating twelve pathology foundation models (PFMs) and ResNet baselines across four different datasets covering diverse, clinically relevant patch-level classification tasks. To our knowledge, this is the first systematic framework that jointly analyses perturbation-level robustness and feature-space domain generalisation in PFMs as follows:

\begin{enumerate}[leftmargin=0pt,labelindent=0pt,itemsep=0.3em]

    \item \textbf{Perturbation-based Robustness Analysis}  
    Models are evaluated under controlled, clinically motivated image perturbations using REET~\cite{reetoolbox}. The perturbations span three categories: pixel-level (noise, brightness, compression), stain and colour-space (stain variation and HED-based mixing), and geometric and structural (blur, rotation, crop, and zoom). This analysis quantifies model resilience across varying perturbation magnitudes and introduces the \textit{Perturbation Performance Index (PPI)} as a summary measure that integrates area under the receiver operating characteristic curve (ROC-AUC) performance over a constraint sweep to capture overall robustness.

    \item \textbf{Non-Redundant K-fold (NR-Kfold) Validation for Domain-Shift Analysis}      
While perturbation analysis captures model sensitivity to controlled, low-level variations, it does not reflect higher-order distribution shifts arising from differences in scanners, staining, or patient populations. To address this, we employ the \textit{Non-Redundant K-fold (NR-Kfold)} protocol, which clusters image embeddings in feature space and assigns entire clusters to distinct folds (see Fig.~\ref{fig:full_benchmark}(b)(i)). Unlike random or stratified splits, this enforces feature-space dissimilarity between training and test sets, inducing controlled covariate and class-conditional shifts in $p(x)$ and $p(x|y)$ that emulate real multi-centre variability~\cite{jahanifar2025domain}. Evaluating models under these deliberately dissimilar folds approximates deployment across sites, providing a direct measure of generalisation robustness under clinically meaningful distribution shifts.
\end{enumerate}

Subsequent sections detail datasets, models, and  robustness–generalisation analyses (Implementation  URL: github.com/DhLYa/pfm-robustness-toolkit).
\subsection{Datasets and Classification Problems}

We use four patch-level histopathology datasets:

\noindent\textbf{National Centre for Tumour Diseases (NCT)}~\cite{NCT_dataset}.  
Contains H\&E patches (224$\times$224, 0.5\,MPP) with Macenko normalisation and nine tissue labels collapsed into a binary task: tumour (TUM) vs non-tumour (ALL; $\sim$1:8 ratio). The NCT-VAL-HE-7K split is patient-disjoint. For computational efficiency, we sample 10\% of the training/validation data and 2.5--10\% for testing, preserving class balance.

\noindent\textbf{Prostate cANcer graDe Assessment (PANDA)}~\cite{PANDA}.  
Comprises multi-centre Gleason-graded WSIs. Pure grades (0+0, 3+3, 4+4, 5+5) are retained, from which 512$\times$512 patches (stride 256) with $\geq$75\% target-class area are extracted. Classes 1--3 are grouped as cancer (positive) vs Class~0 (benign).

\noindent\textbf{PanNuke}~\cite{PanNuke1,PanNuke2}.  
Repurposed for binary tumour detection: positive if a patch contains $\geq$5 neoplastic nuclei and negative if none (excluding 1--4). Predefined three-fold splits are used to ensure tissue diversity.

\noindent\textbf{PatchCamelyon}~\cite{PatchCAM2}.  
Includes $\sim$330{,}000 patches from Camelyon16 lymph-node WSIs labelled for metastasis presence or absence following tissue segmentation. Patient-disjoint, balanced train/validation/test splits ensure unbiased evaluation.

\subsection{Models Evaluated and Linear Probing}
We evaluate two convolutional ResNet baselines and twelve  PFMs spanning the evolution from early transformer encoders to billion-parameter architectures (below). All encoders are frozen, and a single linear classification layer is trained using cross-entropy loss (linear probing) for each task. 

\noindent\textbf{Convolutional baselines.}  
ResNet18 and ResNet50~\cite{ResNet18,ResNet50} serve as standard CNN comparators for assessing the advantages of sophisticated PFMs.

\noindent\textbf{Early transformer PFMs.}  
Hibou-B/L~\cite{hibouB,hibouL}, Phikon-v1/v2~\cite{phikonv1,phikonv2}, EXAONEPath~\cite{exaonepath}, and UNI~\cite{uni} represent mid-sized (80–300M) ViT encoders.

\noindent\textbf{Efficient and medium-scale PFMs.}  
Virchow/Virchow2~\cite{virchow,virchow2} and UNI2~\cite{uni2} lie in the 600–700M range with strong transferability.

\noindent\textbf{Large-scale PFMs.}  
GigaPath~\cite{gigapath} and H-Optimus-0/1~\cite{hoptimus0,hoptimus1} exceed one billion parameters and define the upper limit for capacity scaling in current PFMs.

\subsection{REET Perturbation Analysis with PPI}

Model robustness to controlled perturbations was evaluated using the REET framework~\cite{reetoolbox}.  
Each perturbation is expressed as a parametric transformation $T(x;\boldsymbol{\phi})$ with parameters $\boldsymbol{\phi}$ (e.g., rotation angle, blur radius, stain coefficients).  
A \textit{constraint budget} $\epsilon$ limits deviation $d(\cdot,\cdot)$ from nominal or identity transform parameters $\boldsymbol{\phi}_0$ within  
$\mathcal{C}(\epsilon)=\{\boldsymbol{\phi}:d(\boldsymbol{\phi},\boldsymbol{\phi}_0)\le\epsilon\}$.  
For a fixed pre-trained model $f(\cdot;\boldsymbol{\theta}^*)$ and image patch example $x$ from a dataset with true label $y$, REET finds the perturbation that maximises model loss under this constraint: $
\max_{\boldsymbol{\phi}\in\mathcal{C}(\epsilon)}\;
\mathcal{L}\!\left(f(T(x;\boldsymbol{\phi});\boldsymbol{\theta}^*),y\right),
$
producing an optimised perturbed image $x'_{\epsilon}=T(x;\boldsymbol{\phi}^*)$.  
Optimisation is gradient-based for differentiable perturbations (e.g., pixel noise, adversarial stain) and stochastic for non-differentiable ones (e.g., rotation, cropping, compression); see~\cite{reetoolbox} for details.

\begin{figure*}[!t]
  \centering

  \begin{subfigure}[t]{0.80\textwidth}
    \centering
    \includegraphics[width=\textwidth]{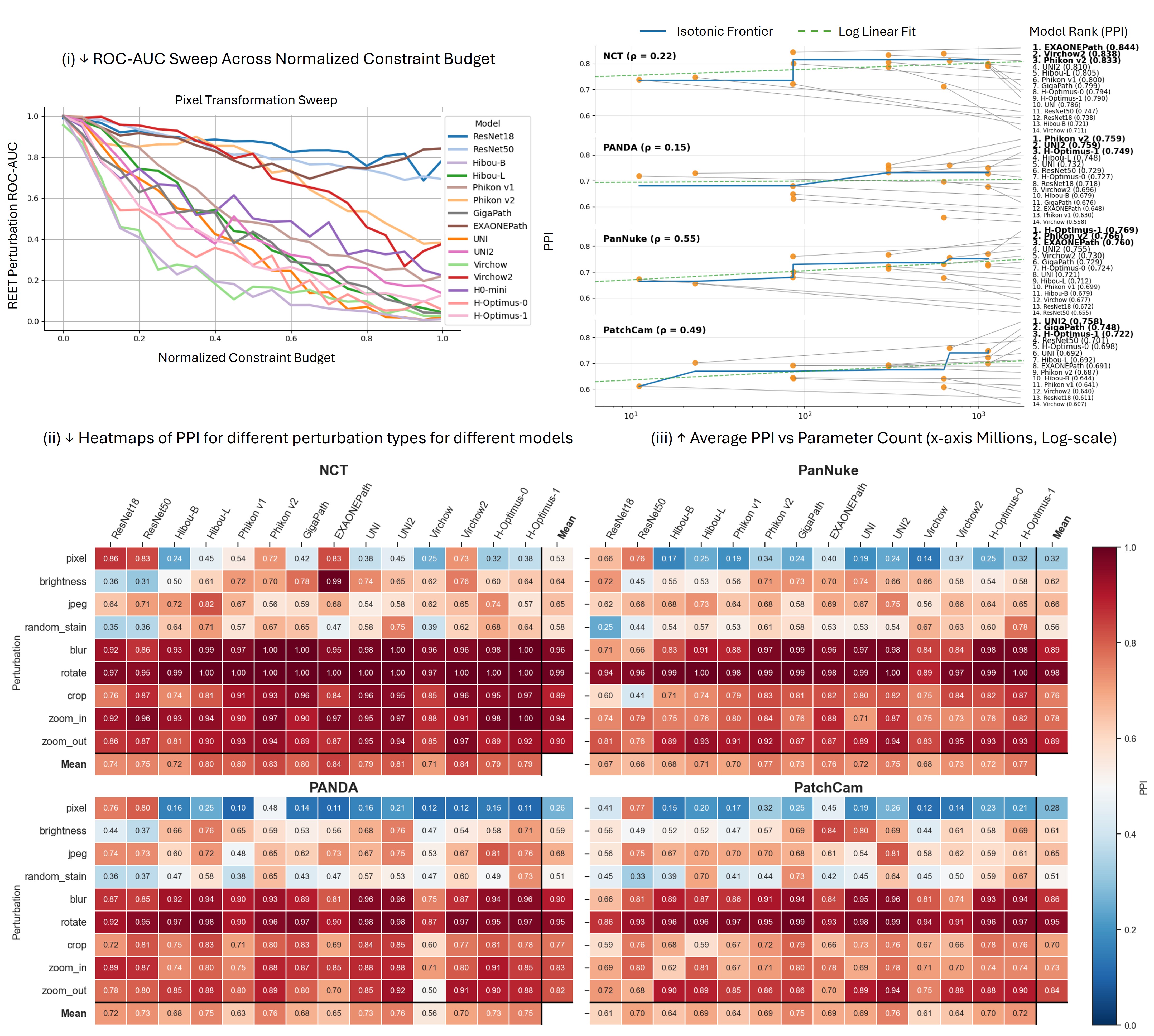}
    \caption{\textbf{Perturbation analysis Results.} 
    \textbf{(i)} ROC–AUC versus normalised constraint sweep for pixel noise; PPI quantifies robustness as the area under this curve.. 
    \textbf{(ii)} PPI heatmaps across datasets, models, and perturbations (higher is better).
    \textbf{(iii)} PPI vs model parameter counts for different datasets.}   
    \label{fig:ppi_analysis}
  \end{subfigure}

  \vspace{0.8em}

  \begin{subfigure}[t]{0.80\textwidth}
    \centering
    \includegraphics[width=\textwidth]{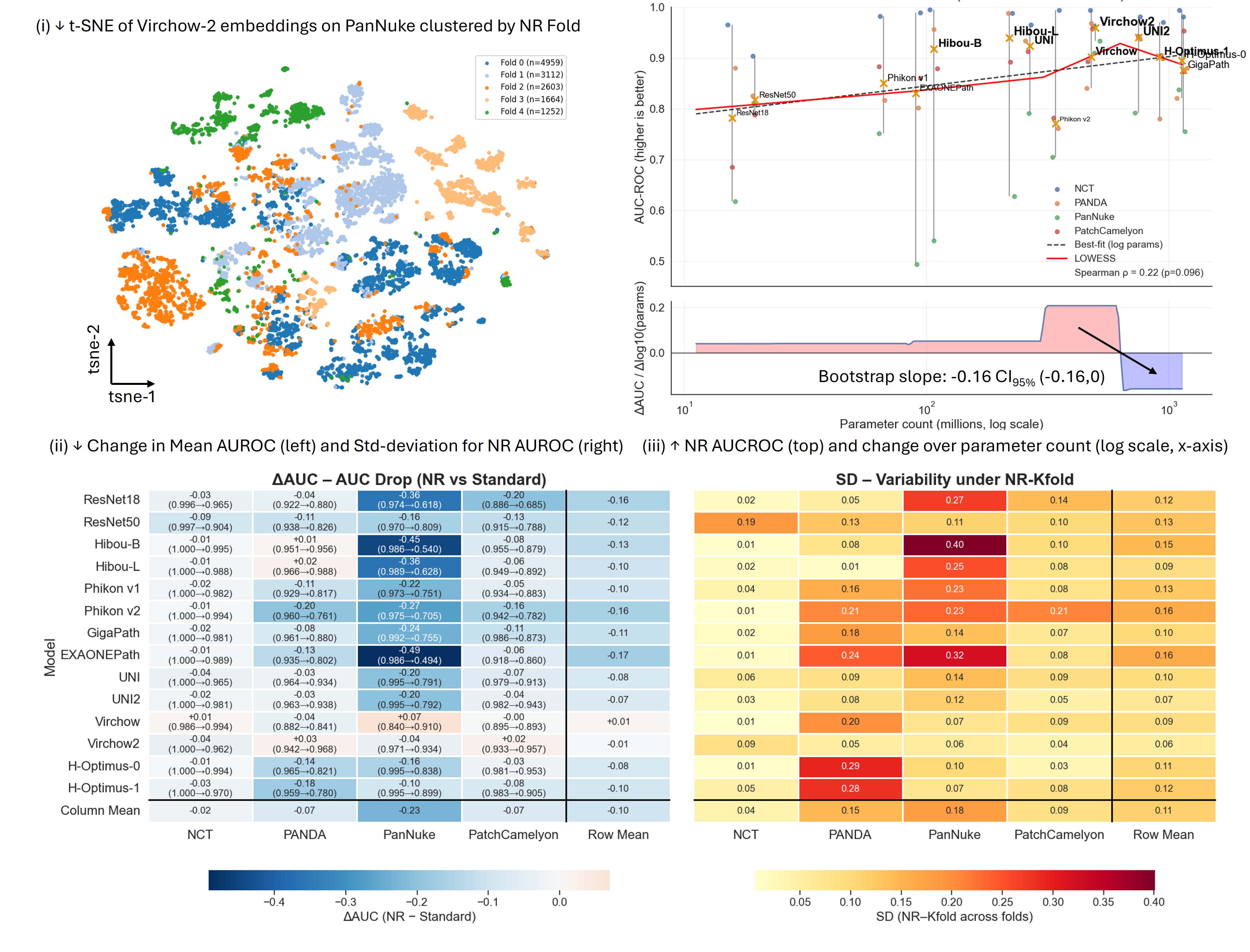}
    \caption{\textbf{Non-Redundant (NR) validation analysis.} 
    \textbf{(i)} t-SNE visualisation of Virchow-2 embeddings on PanNuke, illustrating construction of NR/dissimilar folds by clustering feature embeddings;    
    \textbf{(ii)} Change in mean ROC–AUC (left) and fold-wise standard deviation (right) when moving from standard to NR validation across models and datasets.
    \textbf{(iii)} NR-Kfold ROC–AUC versus model parameter count (top) and corresponding trend in $\Delta$AUC per unit change in $\log_{10}$ parameters (bottom).}
    \label{fig:nr_kfold}
  \end{subfigure}

  \caption{\textbf{Comprehensive robustness and generalisation benchmarking.}}
  \label{fig:full_benchmark}
\end{figure*}

To summarise robustness, we define a \textit{Perturbation Performance Index} (PPI) for each model $m$ and transformation $T$ as the \emph{area under the sweep curve} of test ROC–AUC versus normalised perturbation constraint budget (see Fig.~\ref{fig:ppi_analysis}(i) for an example sweep of pixel perturbations of different strengths over examples from NCT datasets showing how AUC-ROC of different models changes under such perturbations). The robustness of each model $m$ under transformation $T$ is evaluated on a test set of perturbed samples from each dataset  
$\{x'_{\epsilon}\}$ generated by REET at varying constraint budgets.  
To make constraint budgets comparable across perturbations, the range 
$[\epsilon_{\min}^{T},\epsilon_{\max}^{T}]$ is linearly normalised to $\tau\!\in\![0,1]$, 
with $\mathrm{AUC}_{m}^{(T)}(\tau)$ denoting the test ROC–AUC at perturbation level $\tau$.  
The \textit{Perturbation Performance Index (PPI)} is defined as
\begin{equation}
\mathrm{PPI}_{m}^{(T)} \;=\; \int_{0}^{1} \mathrm{AUC}_{m}^{(T)}(\tau)\, d\tau 
\;\approx\; \frac{1}{L}\sum_{l=1}^{L}\mathrm{AUC}_{m}^{(T)}(\tau_l),
\end{equation}
where $\{\tau_l\}_{l=1}^{L}$ are uniformly spaced constraint levels.  
Hence, $\mathrm{PPI}$ represents the average model performance across the entire perturbation spectrum, 
with $\mathrm{PPI}=1$ indicating perfect robustness and smaller values reflecting greater degradation.

\clearpage
\clearpage

\subsection{NR-Kfold Distribution-Shift Evaluation}

To evaluate out-of-domain generalisation, each dataset was partitioned into five \textit{non-redundant folds} by clustering feature embeddings using cosine $k$-means ($k=5$) within each class and assigning entire clusters to distinct folds simulating various domain shifts as discussed earlier  (see Fig.~\ref{fig:full_benchmark}b(i) for an example using Virchow-2 over PanNuke where folds occupy different areas of the feature space). A linear probe was trained on four folds (with a stratified 30\% validation subset) and tested on the held-out fold; weights were reinitialised at each iteration. The process was repeated across all folds, and mean performance was compared with that from standard (random) K-fold stratified cross-validation. The resulting difference, $\Delta$AUC$_\text{NR–CV}$, quantifies the change in performance under deliberate distribution shifts and serves as a dataset-specific measure of each model’s generalisation robustness.

\section{Results and Discussion}


\subsection{Perturbation based Robustness}
As shown in Fig.~\ref{fig:ppi_analysis}(ii), robustness patterns were structured by both perturbation category and model family. 
Pixel-level distortions (noise, brightness, JPEG compression) remained the most destabilising, producing the lowest PPIs ($\approx$0.4--0.6), whereas stain and colour-space variations induced moderate degradation ($\approx$0.6--0.75). 
Geometric perturbations (blur, rotation, crop, zoom) were best tolerated ($\approx$0.8$-$1.0), reflecting strong spatial invariance in ViT-based PFMs. 
Across models, EXAONEPath, Virchow2, UNI2, and Phikon~v2 consistently ranked highest, while smaller PFMs (Hibou-B/L, Phikon~v1) and ResNets underperformed, particularly under pixel or stain stress. 
These results confirm that robustness now depends primarily on pre-training diversity and architectural design rather than parameter count alone.

Figure~\ref{fig:ppi_analysis}(iii) illustrates how robustness, measured by mean PPI, varies with parameter count across datasets. 
Despite spanning over three orders of magnitude in scale, correlations between parameter count and robustness remain weak ($\rho = 0.15$--$0.55$). 
The isotonic frontier indicates a modest improvement from small to mid-sized PFMs, followed by clear saturation beyond approximately 300-600 million parameters. 
Across datasets, mid-sized models such as EXAONEPath, Phikon~v2, Virchow2, and UNI2 consistently occupy the upper frontier, while billion-parameter architectures (H-Optimus-0/1, GigaPath) yield no further gain and occasionally decline. 
These findings confirm that architectural design and domain-specific pre-training rather than sheer parameter count govern robustness performance. 
Overall, robustness gains now depend more on data quality, pre-training diversity, and cross-domain alignment than on further model expansion.

\subsection{Non-redundant testing robustness}
Evaluation under the Non-Redundant K-fold protocol revealed that all models experience measurable degradation when training–test similarity is deliberately broken (Fig.~\ref{fig:full_benchmark}b(ii)). Mean performance loss ($\Delta$AUC) ranged from –0.02 on NCT to –0.23 on PanNuke, reflecting its non-independent train–test composition with patient and WSI overlap. Despite this, pathology foundation models (PFMs) remained markedly more resilient than convolutional baselines. Large and mid-sized PFMs such as Virchow2, UNI2, and H-Optimus-0/1 sustained only modest accuracy loss (mean $\Delta$AUC$\approx$–0.07) and low fold-wise variability (SD$\approx$0.09–0.12), indicating stable generalisation across dissimilar data domains. In contrast, earlier or smaller transformers (e.g., Phikon, Hibou) and ResNet baselines showed larger drops (up to –0.36 AUC) and higher instability (SD$>$0.15), highlighting vulnerability to dataset shifts.

Aggregating NR-Kfold results across datasets further revealed that scaling model parameters does not consistently improve out-of-domain accuracy (Fig.~\ref{fig:full_benchmark}b(iii)). The global correlation between parameter count and AUC-ROC was weak ($\rho=0.22$, $p=0.096$). The Locally Weighted Scatterplot Smoothing (LOWESS) trend showed mild gains up to mid-sized PFMs (Virchow2, UNI2) but clear saturation and slight decline beyond 600M parameters (bootstrap slope was negative $-0.16$, $\mathrm{CI}_{95\%}[-0.16,0]$), indicating that robustness and generalisation have reached a capacity plateau where further scale yields diminishing or even adverse returns. Overall, these findings suggest that breaking latent similarity between training and test distributions exposes hidden brittleness even in high-performing PFMs.

\section{Conclusions}
This study provides the first comprehensive benchmark of PFMs under controlled perturbations and structured distribution shifts. PFMs consistently surpass convolutional baselines in both robustness and domain generalisation, yet performance gains plateau with increasing scale. These results highlight that domain-specific pretraining, data diversity, and representation quality rather than parameter count are the primary drivers of real-world reliability. Future progress should focus on multimodal integration, domain-aligned pretraining, and augmentation strategies that reflect clinical variability to achieve clinically deployable pathology AI systems. Although this analysis is limited to patch-level inference with frozen encoders and a predefined perturbation suite, it demonstrates the importance of systematic robustness evaluation as a standard component of model validation before clinical deployment.

\bibliographystyle{IEEEbib}
\bibliography{references}

\end{document}